\documentclass{article}

\PassOptionsToPackage{numbers, compress}{natbib}

\usepackage[preprint]{neurips_2025}

\usepackage[utf8]{inputenc} 
\usepackage[T1]{fontenc}    
\usepackage{hyperref}       
\usepackage{url}            
\usepackage{booktabs}       
\usepackage{amsfonts}       
\usepackage{nicefrac}       
\usepackage{microtype}      
\usepackage{multicol}
\usepackage{multirow}
\usepackage{makecell}
\usepackage{enumitem}
\usepackage[table]{xcolor}
\usepackage{amssymb}
\usepackage{amsmath}
\usepackage{mathtools}
\usepackage{amsthm}
\usepackage{adjustbox}
\usepackage{wrapfig}
\usepackage{fontawesome}

\title{Understand, Think, and Answer: Advancing Visual Reasoning with Large Multimodal Models}

\author{
Yufei Zhan\textsuperscript{1,2,}$^*$, Hongyin Zhao\textsuperscript{1,}\thanks{Equal Contribution.}\;\;, Yousong Zhu\textsuperscript{1,\faEnvelopeO}, Shurong Zheng\textsuperscript{1,3}, Fan Yang\textsuperscript{1,3},\\\textbf{Ming Tang}\textsuperscript{1,2}, \textbf{Jinqiao Wang}\textsuperscript{1,2,3,4}\\
{$^{1}$ Foundation Model Research Center, Institute of Automation,}\\
{Chinese Academy of Sciences, Beijing, China}\\
{$^{2}$ School of Artificial Intelligence, University of Chinese Academy of Sciences, Beijing, China}\\
{$^3$ Peng Cheng Laboratory, Shenzhen, China}\;\;\;
{$^4$ Wuhan AI Research, Wuhan, China}\\
{\tt\small \{zhanyufei2021, zhaohongyin2020, zhengshurong2023,  yangfan\_2022\}@ia.ac.cn}\\
{\tt\small \{yousong.zhu, tangm, jqwang\}@nlpr.ia.ac.cn}\\
}

\begin{document}

\maketitle

\begin{abstract}
  Large Multimodal Models (LMMs) have recently demonstrated remarkable visual understanding performance on both vision-language and vision-centric tasks. However, they often fall short in integrating advanced, task-specific capabilities for compositional reasoning, which hinders their progress toward truly competent general vision models. To address this, we present a unified visual reasoning mechanism that enables LMMs to solve complicated compositional problems by leveraging their intrinsic capabilities (e.g. grounding and visual understanding capabilities). Different from the previous shortcut learning mechanism, our approach introduces a human-like understanding-thinking-answering process, allowing the model to complete all steps in a single pass forwarding without the need for multiple inferences or external tools. This design bridges the gap between foundational visual capabilities and general question answering, encouraging LMMs to generate faithful and traceable responses for complex visual reasoning. Meanwhile, we curate 334K visual instruction samples covering both general scenes and text-rich scenes and involving multiple foundational visual capabilities. Our trained model, Griffon-R, has the ability of end-to-end automatic understanding, self-thinking, and reasoning answers. Comprehensive experiments show that Griffon-R not only achieves advancing performance on complex visual reasoning benchmarks including VSR and CLEVR, but also enhances multimodal capabilities across various benchmarks like MMBench and ScienceQA. Data, models, and codes will be release at \url{https://github.com/jefferyZhan/Griffon/tree/master/Griffon-R} soon.
\end{abstract}

\begingroup
\renewcommand*{\addcontentsline}[3]{}
\section{Introduction}
Inspired by the success of Large Language Models like ChatGPT \cite{openai2023gpt4} and Gemini \cite{team2023gemini}, the vision field has been seeking to equip these models with visual understanding capabilities, aiming to replicate similar achievements in visual tasks. Currently, Large Multimodal Models (LMMs) \cite{liu2024visual, li2023blip, instructblip, li2024monkey, liusphinx} adopt a paradigm in which images are encoded and projected into a textual embedding space, then combined with language input to generate responses via the LLM \cite{vicuna2023, gao2023llama}. Trained with millions of high-quality data, LMMs demonstrate advancing performance across various vision-language tasks, such as visual question answering (VQA) \cite{vqa} and image captioning \cite{chen2015microsoft}, and become increasingly proficient in fine-grained visual tasks like visual grounding \cite{flickr30k} and object detection \cite{lin2014microsoft}, even surpassing specialized vision expert models \cite{ren2015faster, kamath2021mdetr} in certain domains.
\begin{figure}[t]
  \centering
   \includegraphics[width=0.7\linewidth]{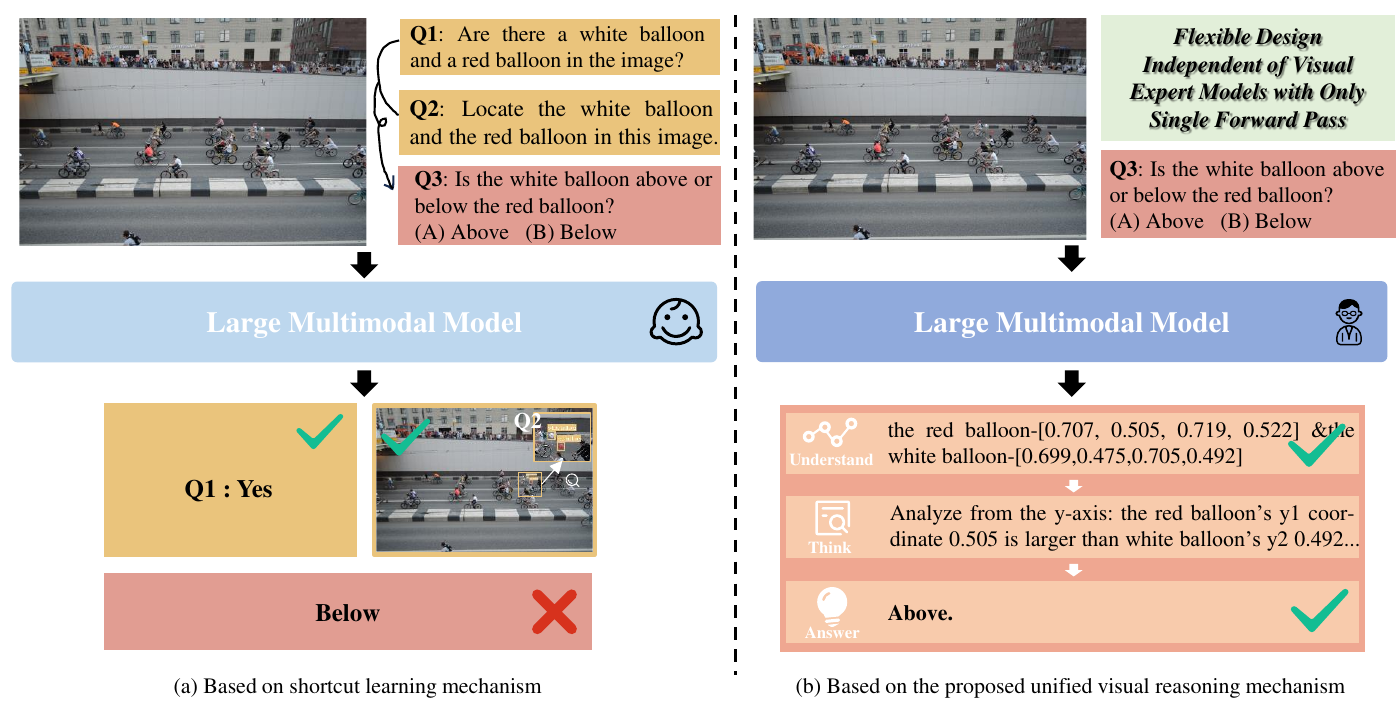}
   \caption{Enabled by the proposed unified mechanism, Griffon-R naturally connects the reasoning processes for locating each balloon with answering the spatial relationship question. It effectively analyzes their y-axis coordinates and provides the correct answer in a single pass.}
   \label{fig:Compare}
\end{figure}

Despite significant progress across a wide range of tasks, LMMs still fall short in visual reasoning tasks. Existing open-source LMMs mainly follow the shortcut learning paradigm\cite{geirhos2020shortcut} and are trained to directly generate the final answer based on the question. As shown in Fig. \ref{fig:Compare}(a), LMMs do well in visual foundational tasks that follow the shortcut paradigm like object recognition(Q1) and visual grounding(Q2). Though this paradigm benefits the LMMs a lot, it also hinders them from inferring based on the foundational visual capabilities, which are crucial for tackling compositional and complicated visual reasoning tasks. As indicated in Fig. \ref{fig:Compare}(a), when directly asking the LMMs about the relative position of the red and white balloons, they respond with the unfaithful but confident answer, \textit{i.e.} hallucination \cite{yin2023woodpecker}.

To address these challenges, recent studies have shown that incorporating multimodal Chain-of-Thought (CoT) \cite{shao2024visualcot, li2024vocot, MitraCCoT} to encourage LMMs to reason step by step can improve the reason quality. However, these CoT methods are specifically designed for target types of questions or domains and may require multiple times of forwarding. Other toolkit-based methods \cite{qi2024cogcom, jain2024vcoder, wu2024v} enhance LMMs by directly supplementing details needed for specific tasks using the visual toolkit with the format of structured text or feature-based prompts. Though the visual toolkit can provide accurate information, calling it brings significant computational load and latency. Also, with the task complexity increasing, the parameters scale up exponentially. 

In this paper, we propose a unified visual reasoning mechanism to enable LMMs to harness advanced intrinsic visual foundational capabilities for compositional visual reasoning in a single forward pass. Inspired by the human reasoning process\cite{anderson2005cognitive,mcvee2005schema}—where individuals begin by contemplating how to answer a question, gather sufficient information from their environment, think with their experiences, and ultimately formulate an answer—our unified mechanism integrates this progressive ``understand-think-answer'' approach. Following this proposed process, the model first plans the necessary information acquisition for answering the question and generates structured instructions to autonomously gather relevant information, ensuring a thorough understanding. Considering the contextual understanding and extensive knowledge base of LLMs, rather than designing question-specific reasoning paths, the model is self-prompted to engage in contextual thinking after obtaining a comprehensive understanding. This design allows for greater flexibility across various question types. Finally, the model generates an answer, marking the conclusion of the visual reasoning process. This entire process operates without manual intervention or the need for external tools, achieving both efficiency and adaptability in a single forward pass. To implement this mechanism, we introduce a semi-automatic expert-supervised data engine and curate a dataset of 334K visual reasoning samples, encompassing both natural and textual scenes. This dataset is annotated progressively by employing AI\cite{qwen2.5, wang2024qwen2} and human experts in a streamlined pipeline. Ultimately, we train the Griffon-R model using this curated data to achieve the unified mechanism.

To validate our design, we conduct comprehensive experiments with Griffon-R across a range of visual reasoning and multimodal benchmarks. The results demonstrate that empowered by our design mechanism, Griffon-R achieves advancing performance on complex visual reasoning tasks VSR\cite{liu2023vsr} and CLEVR\cite{johnson2017clevr} and surpasses advanced LMMs on multimodal tasks including MMbench \cite{mmbench}, ScienceQA \cite{sqa}, \textit{etc.}, highlighting its enhanced general capabilities. Our key contributions are as follows:
\begin{itemize}
    \item We propose a unified visual reasoning mechanism inspired by the human reasoning process, enabling LMMs to handle diverse compositional tasks by leveraging advanced capabilities through an "understand-think-answer" process in a single forward pass.
    \item We curate 334K multi-scene visual reasoning data by the introduced semi-automatic expert-supervised data engine and further present Griffon-R, a general LMM that is skilled in solving complicated compositional problems.
    \item We conduct extensive experiments on a wide range of visual reasoning and multimodal benchmarks. Griffon-R achieves advancing performance in compositional VSR and TallyQA, while further boosting performance on the multimodal benchmark including MMBench, ScienceQA, \textit{etc.}.
\end{itemize}

\section{Related Works}

\subsection{Multimodal Chain of Thought}
The CoT approach \cite{wei2022chain, kojima2022large,zhang2022automatic, yao2024tree, besta2024graph} is a series of prompting techniques that improve the ability of LLMs to solve complex reasoning tasks. With the rise of LMMs, these methods have gradually been incorporated into the multimodal domain to enhance model performance on complex reasoning tasks. ScienceQA \cite{sqa} pioneeringly proposes Multimodal CoT to combine image captioning for reasoning, thus enabling models to handle complex question-answering tasks in science. Later works further decompose complex visual reasoning tasks into sequential steps, incorporating diverse prompts such as bounding boxes \cite{chen2023shikra, li2024vocot, shao2024visualcot}, textual descriptions \cite{zheng2023ddcot}, and scene graphs\cite{MitraCCoT}, allowing models to reduce intuitive errors by following a structured reasoning path. However, these multimodal CoT methods usually design specific paths for considered tasks and thereby limiting the model’s ability to generalize across diverse question types and visual tasks. In this way, methods like Visual CoT\cite{shao2024visualcot} tend to provide rough prompts when questions are not included in these designed patterns. Also, some of them\cite{shao2024visualcot, li2024vocot} re-forward regions during inference to enhance region understanding. In contrast, our approach keeps the streamlined model structure and inference process with a single forward pass, which is more efficient. Meanwhile, our mechanism analyzes the question to automatically leverage intrinsic capabilities to understand the image precisely, facilitating a more flexible pattern for better generalization.

\subsection{Toolkit-Based Visual Reasoning}
Unlike multimodal CoT methods that convert various prompts into text for guidance, visual-tool-using approaches\cite{lei2024scaffolding,yang2023som,jain2024vcoder,qi2024cogcom, wu2024v,hu2024vpd,gupta2023vp,suris2023vipergpt} directly input information in different formats or modalities into the model, encouraging comprehensive understanding and reasoning. Methods like SoM \cite{yang2023som} and Scaffolding \cite{lei2024scaffolding} initially incorporate additional structured information within images, using these as anchors to prompt GPT-4V in visual reasoning. In contrast to these methods specifically designed for GPT-4V, Vcoder\cite{jain2024vcoder} takes a different approach by projecting depth and segmentation maps into the text embedding space, thereby improving reasoning accuracy by enhancing model comprehension. These methods typically rely on one or two fixed visual tools, limiting the richness of the information provided. Consequently, other approaches \cite{qi2024cogcom, wu2024v,hu2024vpd,gupta2023vp,suris2023vipergpt} have been developed to construct visual toolkits specifically for visual reasoning tasks in LMMs, allowing LLMs to generate executable program calls to specialist visual models based on the question. Building on this, CogCom \cite{qi2024cogcom} further integrates certain visual manipulations with the model’s internal capabilities, enabling it to generate and execute an operation chain to progressively complete reasoning tasks. In comparison to these methods, our method leverages intrinsic capabilities and supports knowledge sharing across contexts, minimizing the risk of errors from isolated task execution. Additionally, with an end-to-end manner, our approach reduces the optimization difficulty and the latency.

\begin{figure}[t]
  \centering
   \includegraphics[width=\linewidth]{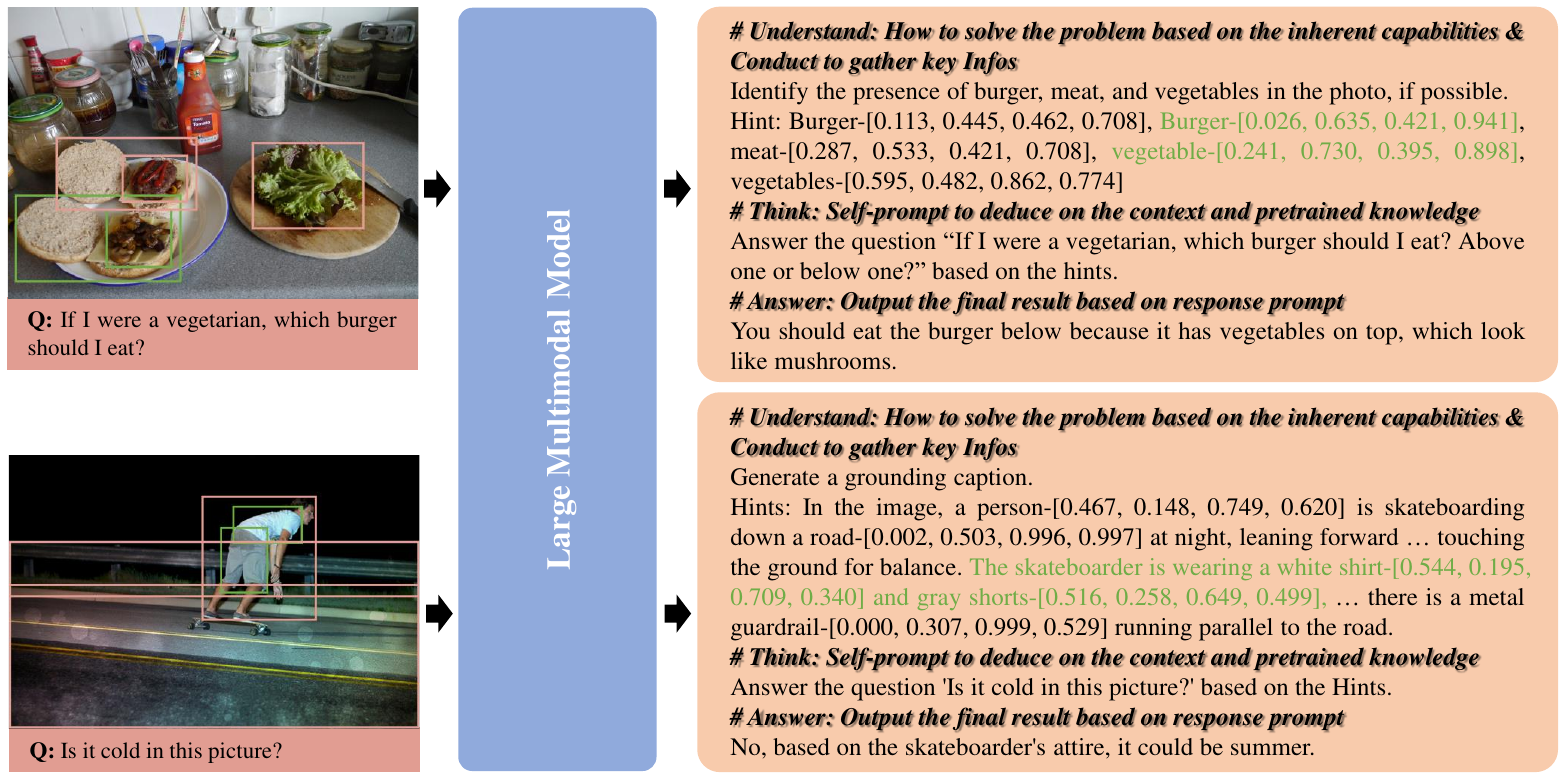}
   \caption{Detailed illustration of the unified visual reasoning mechanism with the ``Understand-Think-Answer'' process. The key information related to the answer is highlighted or visualized with the green color. We illustrate the details of the designed process in bold, which will not be generated or trained.}
   \label{fig:ff}
\end{figure}

\section{Methodology}

In this section, we start with our designed novel unified visual reasoning mechanism, specifically the ``Understand-Think-Answer'' process which bridges the intrinsic visual foundational capabilities and VQA answering to allow the LMMs to reason accurately in a single pass in Sec. \ref{sec: mechanism}. Then, we detail the semi-automatic expert-supervised data engine on how it curates the 334K high-quality visual reasoning data aligned with the mechanism in Sec. \ref{sec: data}. Finally, we present the Griffon-R, an visual reasoning improved LMM built on the proposed unified mechanism and data.

\subsection{Preliminary}

Current LMMs typically adopt an auto-regressive approach to generate response via next token prediction. Specifically, for an LMM $M$ with parameters $\theta$, given an input image $I$ and user instruction $Q$, the model maximizes the sequence probability to output the response sequence $X$:
\begin{equation}
    p(X) = \prod_i p(x_i|I, Q, X_{<i})
\end{equation}
where $X_{<i}$ represents the sequence before the current prediction $x_i$ in the output sequence. Current LMMs employ a shortcut learning paradigm, training the model to directly generate the final answer $X_{ans}$. When simplifying the probability calculation process, the process can be denoted as:
\begin{equation}
    X_{ans} = M_\theta(I, Q)
\end{equation}
Though it effectively solves straightforward question-answering tasks, it struggles with more challenging compositional visual reasoning tasks, as illustrated in Fig. \ref{fig:Compare}. Therefore, we propose the followed unified visual reasoning paradigm that advances the model to perform compositional reasoning leveraging intrinsic capabilities in a single forward pass through ``understand-think-answer'' process.
\subsection{Unified Visual Reasoning Mechanism}
\label{sec: mechanism}

Current LLMs\cite{team2024gemma,qwen2.5} trained on billions of data have acquired rich knowledge and experience, while LMMs through image-text alignment\cite{li2023blip, radford2021clip} and instruction tuning\cite{instructblip, liu2024visual} further enhance their ability to perceive and interpret information by themselves. However, they still struggle to imitate how an educated person answers compositional questions by problem analyzing with past experience, relevant information gathering, and reaching a conclusion from thinking\cite{anderson2005cognitive, mcvee2005schema}. Therefore, we propose the unified visual reasoning paradigm to imitate this approach to enable LMMs to sequentially and continuously perform the ``understand-think-answer'' process, accurately arriving at the answer in a single pass without introducing any toolkit.

\textbf{Understand.} Understanding is a key step that directly impacts the accuracy of the response. Unlike direct answering, the models first analyze the question and image to determine how to approach the problem and which intrinsic abilities to use to extract relevant information. Based on this analysis, the model employs the appropriate capabilities to gather key information, achieving a thorough understanding of the image in relation to the question. As shown in Fig. \ref{fig:ff}, we combine question analysis with intrinsic capability-based information retrieval planning and generate instructions for acquiring the required information. Then, they automatically guide the model to gather relevant visual cues. For the above sample in Fig. \ref{fig:ff}, the model identifies that solving the question requires knowing the location of the burger and distinguishing which items contain meat or vegetables, as indicated in the instruction by the terms ``burger,'' ``meat,'' and ``vegetables''. The ``Indentify the presense'' process involves applying these capabilities to gather the necessary information. This understanding process leverages common capabilities of LMMs, including caption, grounded caption, visual grounding, text recognition, \textit{etc.}, instead of relying on any fixed capability like scene analysis\cite{MitraCCoT} or REC\cite{shao2024visualcot}. Moreover, when no relevant information is found, the model outputs none instead of providing vague clues\cite{shao2024visualcot, wu2024v}, avoiding potential misleading.

\textbf{Think \& Answer.} Instead of designing a specific reasoning path, we adopt a self-prompt approach. Previous studies show current models\cite{chatgpt2024} can effectively respond to questions based on contextual cues. Additionally, modern LMMs\cite{you2023ferret, yuan2024osprey} excel at understanding coordinates-format object references in the context without needing additional forward passes\cite{li2024vocot,shao2024visualcot} for object perception. After achieving a deep understanding, we allow the model to generate the instruction to encourage the model to engage in self-thinking based on the visual cues, as indicated in Fig. \ref{fig:ff}. Ultimately, the model generates the final reasoning output according to the response template in the user’s instruction. By now, the model performs question-customized deep understanding of the image in a single forward pass, engages in self-prompted thinking based on the generated cues, and outputs the final answer. This process can be summarized as:
\begin{gather*}
    p([X_U, X_T, X_{ans}]) = \prod_i p(x_i|I, Q, X_{<i})\\
    X_{ans} = M_\theta(I, Q, X_U, X_T)
\end{gather*}
where $X_U$ denotes the context generated during the understanding and $X_T$ denotes the context generated during the thinking. The sequence inside the square brackets [] is generated with a left-to-right order of generation. Compared to shortcut learning-based methods, our unified visual reasoning mechanism integrates intrinsic capabilities to provide more accurate solutions to combinatorial visual reasoning tasks. In contrast to other CoT and toolkit-based approaches, it offers both flexibility and efficiency, which eliminates the need for multiple forward passes.

\subsection{Expert-Supervised Data Engine}
\label{sec: data}

\begin{wrapfigure}[12]{r}{0.5\textwidth}
    \centering
    \includegraphics[width=0.95\linewidth]{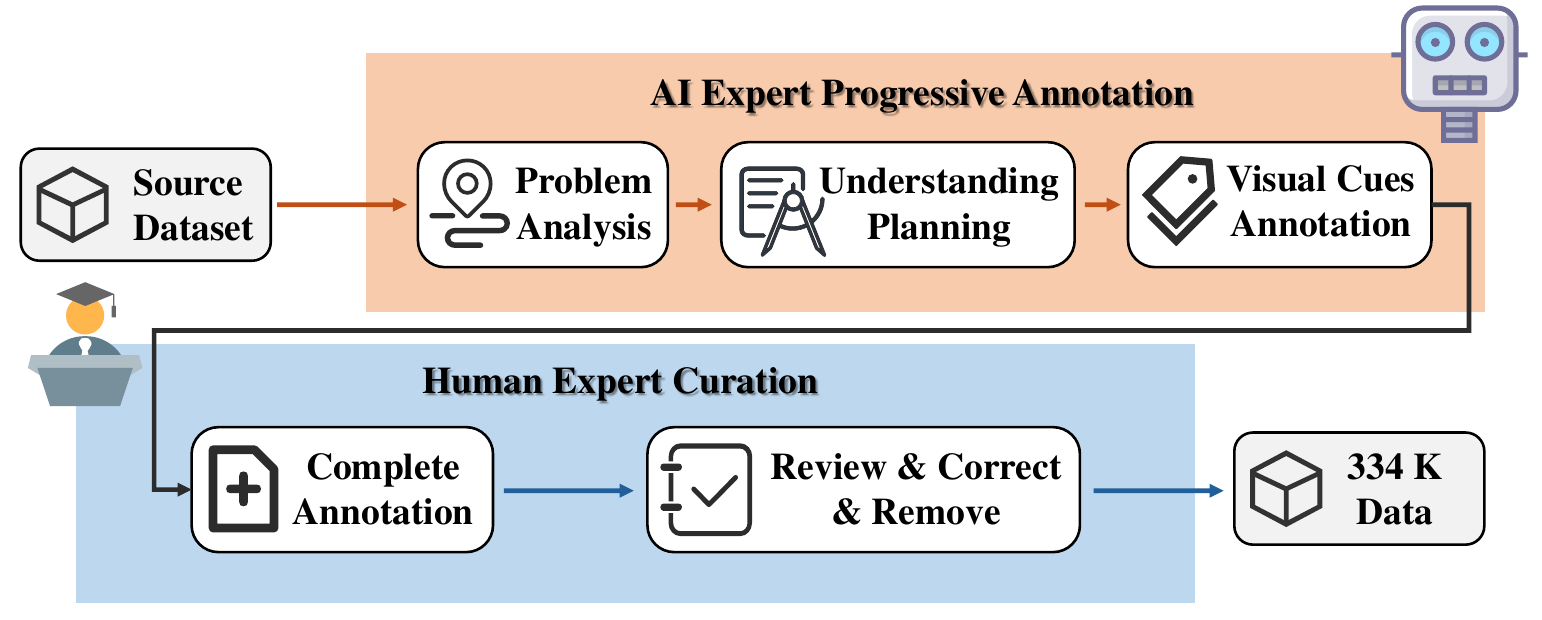}
    \caption{Illustration of the semi-automatic expert-supervised data generation engine.}
    \label{fig:data}
\end{wrapfigure}

Beyond the design of the mechanism, training LMMs for accurate visual reasoning remains a significant challenge due to data limitations. To address this challenge and support the implementation of our designed visual reasoning mechanism, we introduce a semi-automatic expert-supervised annotation engine in this section, which is designed to generate high-quality visual reasoning data with visual cue annotations following our mechanism. As shown in Fig. \ref{fig:data}, this semi-automated approach, supported by experts, enables us to generate high-quality data that enhances the model’s visual reasoning capabilities through the unified reasoning mechanism.

\textbf{Progressive Annotation with AI Expert.} 
Due to the advanced question-answering capabilities and vast knowledge of current LMMs trained with billions of tokens, we leverage state-of-the-art LMMs to assist with annotation based on our visual reasoning mechanism. We select the Qwen2-VL-72B model\cite{wang2024qwen2} from the latest open-source models as our AI Expert, due to its strong task performance. For each sample, the AI Expert is responsible for designing the understanding process for solving problems and annotating tasks it excels in. First, we prompt the AI Expert to analyze the question and image to determine how to solve the problem and identify the necessary information. Then, we further instruct the expert to outlining the understanding process with specific tasks to gather key information based on the common intrinsic capabilities of LMMs. Finally, for tasks that the AI Expert specializes in, we directly use the model to generate the corresponding annotations, such as caption, text recognition, \textit{etc}. The instructions used in this process are described in the Appendix \ref{app: data}.

\textbf{Curation with Human Expert.} 
Although the AI Expert is skilled in analyzing and annotating tasks related to the question, its capabilities are limited in complex visual reasoning and tasks, such as visual grounding involving multiple objects. Previous works\cite{chen2023sharegpt4v, yuan2024osprey} have also highlighted that human-involved high-quality annotations form the foundation for scaling data size in subsequent phases. Therefore, after the AI Expert’s annotation, we first employ human experts to complete tasks that the AI Expert is not proficient at, primarily visual grounding in scenes with multiple objects or partial text. Then, human experts further review the annotations for quality, including evaluating whether the understanding process is sound, ensuring the accuracy of the task annotations, and eliminating overly simplistic questions.

\textbf{Data Description.} Leveraging our proposed semi-automatic expert-supervised data generation pipeline, we curate 334K image-question pairs from multi-task instruction-following data across multiple levels from public datasets. We mainly focus on the general scene data and text-rich scene data, which covers compositional reasoning problems. The data annotations are then completed through the above semi-automatic expert-supervised annotation process, which incorporates the AI expert progressive annotation stage and human expert curation stages. We provide more information about the source data in the Appendix \ref{app: data}.

\subsection{Griffon-R}
\label{sec: model}
Leveraging the proposed unified visual reasoning mechanism and the curated 334K data, we further develop the Griffon-R model, an LMM proficient at both compositional visual reasoning tasks and straightforward QA tasks. Benefiting from our design, Griffon-R maintains a streamlined architecture without the need for additional perception structures. Also, Griffon-R generates responses only through the next token prediction, rather than relying on module calling or multiple forward passes. We detail the construction process, including the architecture and training pipeline, which may also serve as a guideline to facilitate the implementation of the unified visual reasoning mechanism.

\textbf{Structure.} Griffon-R adopts the advanced single-branch high-resolution structure proposed in Griffon v2 \cite{zhan2023griffon}. Compared with other LMMs, this structure is proved to be better at fine-grained object localization, which is important for most of visual reasoning tasks. It consists of three core components: a high-resolution visual encoder, a vision-language connector, and an LLM. The high-resolution visual encoder processes image inputs up to 1K resolution without partitioning, with the connector projecting and compressing the tokens while retaining the performance.

\textbf{Training Pipeline.} We follow the common practice \cite{li2024vocot, shao2024visualcot} to train the Griffon-R model. Specifically, after the basic pertaining stage, we combine the curated visual reasoning data following our mechanism with VQA and instruction data to fine-tune the model with the whole model updated. Differently, we only use the cross-entropy loss to supervise the training without introducing task-specific losses\cite{wu2024v}. We detail the training data and setting for each stage in the Appendix \ref{app: train}.

\section{Experiments}

\begin{table}[t]
\centering
\small
\caption{Evaluation results on visual reasoning benchmarks. CLEVR focuses on generated structured scenes, while the others evaluate models in the natural scenes. We use the Spatial subset of V-Star here due to the reasoning step included, while the attribute subset primarily evaluates models' attribute perception abilities in high-resolution scenes.}
\label{table:vr}
\setlength{\tabcolsep}{10pt}
\begin{tabular}{@{}l|cccccc}
\toprule
\multirow{2}{*}{Methods}  &\multirow{2}{*}{VSR} & \multirow{2}{*}{CLEVR} & \multirow{2}{*}{GQA} &\multirow{2}{*}{V-Star$_{Spat.}$} &  \multicolumn{2}{c}{TallyQA} \\ 
&&&&&Simple & Complex\\
\midrule
\multicolumn{7}{c}{Large Multimodal Models}\\
\midrule
BLIP-2-7B \cite{li2023blip}  &50.9 & - & 44.7 &  53.9 & - & - \\
InstructBLIP-7B \cite{instructblip}  & 52.1 & - & 48.3 &  47.4&74.3  &48.7  \\
LLaVA-1.5-7B \cite{liu2023improvedllava}  & 64.2 & 43.7 & 62.0 &  53.9  & 75.9&63.8 \\
LLaVA-1.5-13B \cite{liu2023improvedllava}  & 70.4 & 55.8 & 63.3 &  55.3  &76.9 &65.4 \\
Monkey-7B \cite{li2024monkey}  & 62.9 & 46.3 & 60.7 & 53.9 &80.9  &63.0 \\
DeepSeek-VL-7B \cite{lu2024deepseek}&\underline{67.5} &  48.8 & 61.31 &  40.3  &79.5 &62.1 \\
LLaVA-Next-7B \cite{liu2024llavanext} & 63.8 & 51.9 & 64.2 & 63.2  &80.3 &66.5\\
Ferret v2-7B \cite{you2024ferret} & - & - & 64.7 &  - &  - &-  \\
\midrule
\multicolumn{7}{c}{Large Multimodal Models with MCoTs or Toolkit-Based Enhancement}\\
\midrule
VolCano-7B \cite{li2024vocot} & 67.2 & \underline{56.2} & 64.4 &  50.3  & 73.5 &55.2\\
CogCom-17B \cite{qi2024cogcom}& - & - & \textbf{71.7} & - & \underline{84.0} & 70.1 \\
SEAL-7B \cite{wu2024v}  &48.5& -&  - &  \underline{76.3} & 51.9 &20.6 \\ 
VPD-5B \cite{hu2024vpd}&- &-  &61.3 & -   &83.1 &\textbf{70.9} \\
VisualCoT-7B \cite{shao2024visualcot} & 61.4 &55.5 & 63.1 &  50.3 &  82.4 &60.2 \\
VisualCoT-13B \cite{shao2024visualcot} & - &55.8 & 63.3 &  54.9 &  83.1 &70.3 \\
\midrule
\multicolumn{7}{c}{Large Multimodal Models Leveraging Intrinsic Capabilities}\\
\midrule
Griffon-R-9B  &\textbf{70.9} & \textbf{63.7} & \underline{65.1} &  \textbf{77.6} & \textbf{84.4} &\underline{70.4} \\
\bottomrule
\end{tabular}

\end{table}

\subsection{Implementation Details}
We follow the Griffon v2 \cite{zhan2023griffon} to set the resolution to 1022, randomly initialize a down-sampling projector implemented as a 3×3 convolution (stride 2, padding 1).Then, we utilize CLIP-ViT-L / 14-336 \cite{radford2021clip} to initialize the visual encoder and further interpolate to the defined resolution. To ensure that our framework learns visual language reasoning capabilities from the scratch - without sacrificing generality - we selected Griffon-G-9B architectural paradigm and use Gemma9B \cite{team2024gemma} to initialize the LLM. For the training, we utilize the AdamW optimizer \cite{adam}, setting the learning rate to 1e-3 for the first stage and 2e-5 for stage 2 and stage 3. We use the DeepSpeed zero2 \cite{rasley2020deepspeed} and a cosine learning rate strategy \cite{cosine} with a warmup \cite{warmup} ratio of 0.3. We train each stage for 1 epoch with the batch size of 256.All training was performed on eight NVIDIA H100 GPUs.

\subsection{Evaluation Details}
To comprehensively evaluate Griffon-R’s capabilities, we conduct a fair comparison with advancing LMMs and visual reasoning methods on visual reasoning benchmarks across both structured scenes like CLEVR\cite{johnson2017clevr} and natural scenes, encompassing VSR\cite{liu2023vsr}, GQA\cite{gqa}, TallyQA\cite{acharya2019tallyqa}, and V-Star$_{Spat.}$. These benchmarks mainly require models to comprehend the multi-level information based on the image to infer the final answer. We also evaluate Griffon-R on multimodal benchmarks to demonstrate its comprehensive capabilities. We include benchmarks that contain partial visual reasoning with common sense and knowledge, like MMBench\cite{mmbench}, ScienceQA\cite{sqa}, SEED\cite{li2023seed}, and LLaVA Bench\cite{liu2024visual}, and also TextVQA\cite{textvqa}, which focuses on text-scene understanding. Given that strong visual reasoning skills help mitigate model hallucinations, we also assess performance on the POPE\cite{pope} benchmark. Also, we validate our design specifically in the ablation studies, including understanding quality indicated by grounding task, mechanism, and data. 

\subsection{Evaluation on Compositional Visual Reasoning}
As shown in Tab. \ref{table:vr}, Griffon-R achieves outstanding performance across these visual reasoning benchmarks. It reaches advancing levels in both generated scenes task CLEVR and natural scenes including VSR, V-Star$_{Spat.}$, and TallyQA$_Simple$, outperforming advanced visual reasoning models like Visual CoT and VPD. Specifically, Griffon-R achieves 63.7\% accuracy on CLEVR and 70.9\% accuracy on the VSR, surpassing the second-place models by a large margin. Such remarkable performance demonstrates the strong visual reasoning capabilities of Griffon-R across multiple tasks and scenarios. Notably, it also surpasses the latest advanced LMMs LLaVA-Next and Ferret v2. On the high-resolution, small-object compositional reasoning benchmark V-Star whose scenes are quite challenging for LMMs, Griffon-R outperforms multiple methods based on CoT and visual search techniques. These achievements across diverse task focus further highlight Griffon-R's robust visual reasoning capabilities and validate the effectiveness of our mechanism and data design.

\begin{table}
\centering
\small
\caption{Evaluation results on multimodal benchmarks. Compared with visual reasoning benchmarks, these benchmarks focus more on visual understanding with common sense and knowledge, yet incorporating simple reasoning.}
\label{table:mm}
\setlength{\tabcolsep}{7.5pt}
\begin{tabular}{@{}l|cccccc}
\toprule
{Methods} & MMB & ScienceQA & TextVQA &SEED-Img & LLaVA-W &POPE\\
\midrule
\multicolumn{7}{c}{Large Multimodal Models}\\
\midrule
InstructBLIP-7B \cite{instructblip} & 36.0 & - & 50.1 & 58.8 & 60.9 & 72.1\\
LLaVA-1.5-13B \cite{liu2023improvedllava} & 67.7 & 71.6 & 61.3 & 68.2 & 72.5 & 85.9\\
QwenVL-Chat-7B \cite{Qwen-VL}& 60.6 & 68.2 & 61.5 & 65.4 & - & 84.7\\
Monkey-7B \cite{li2024monkey}& 61.9 & 69.4 & 67.6 & 67.6 & 53.9 & 82.6\\
DeepSeek-VL-7B \cite{lu2024deepseek}& \underline{71.3} & - & 63.7 & \underline{70.4} & 21.6 & 85.8\\
LLaVA-NeXT-7B \cite{liu2024llavanext}& 69.0 & 73.2 & 64.9 & 70.2 & 72.3 & 86.4 \\
Ferret v2-13B \cite{you2024ferret}& - & - & 62.2 & 61.7 & 69.9 & {88.1}\\
\midrule
\multicolumn{7}{c}{Large Multimodal Models with MCoTs or Toolkit-Based Enhancement}\\
\midrule
VolCano-7B \cite{li2024vocot}& 68.1 & 38.3 & 57.4 & 64.5 & 56.5 & 86.5\\
SEAL-7B \cite{wu2024v}& 33.1& -& - & 41.7 & 59.1 & 82.4\\
CCoT-13B \cite{MitraCCoT}& 70.7 & 69.7 &- & 69.7 & \underline{74.9} & -\\
VPD-5B \cite{hu2024vpd}& 69.0 & 83.1 & \underline{70.9} & - & - & \underline{88.6}\\
CogCom-17B \cite{qi2024cogcom}& - & \underline{84.0} & - & - & - & 87.8\\
VisualCoT-7B \cite{shao2024visualcot} & 67.3 & 68.3 & 61.0 & - & 49.7 & 86.5\\
VisualCoT-13B \cite{shao2024visualcot} & 67.4 & 73.6 & 62.3 & - & 57.7 & 83.3\\
\midrule
\multicolumn{7}{c}{Large Multimodal Models Leveraging Intrinsic Capabilities}\\
\midrule
Griffon-R-9B & \textbf{79.0} & \textbf{87.0} & \textbf{72.4} & \textbf{73.8} & \textbf{76.2} & \textbf{89.3}\\
\bottomrule
\end{tabular}
\end{table}

\subsection{Evaluation on Multimodal Benchmarks}
Empowered by the unified visual reasoning mechanism, Griffon-R leverages the model's inherent capabilities for visual reasoning tasks. Meanwhile, Griffon-R with streamlined structure and joint optimization with straightforward QA data can also handle these widely applied multimodal benchmarks. Therefore, we also evaluate Griffon-R on multimodal benchmarks and various VQA tasks to verify its comprehensive capabilities. As shown in Tab. \ref{table:mm}, Griffon-R outperforms representative LMMs as well as advanced visual-reasoning-enhanced LMMs. Griffon-R achieves a score of 79.0 on the comprehensive MMBench and excels on ScienceQA, which focuses on scientific reasoning. Also, in real scenarios, Griffon-R is more proficient at challenging tasks like SEED and LLaVA-in-the-wild benchmark. In text-based scenes, Griffon-R further surpasses the program-driven VPD model, achieving 72.4\% accuracy on TextVQA. These evaluations demonstrate that Griffon-R not only precisely handles complex compositional reasoning tasks but also performs well in general visual question answering.

\subsection{Ablation Studies}

\begin{wrapfigure}[15]{r}{0.5\textwidth}
    \vspace{-\baselineskip}
    \centering
    \caption{Ablation on understanding quality. With REC task covering object localization and attribute perception, we choose it to evaluate the quality of understanding in the mechanism.}
    \label{table: rec}
    \adjustbox{max width=\linewidth}{
    \setlength{\tabcolsep}{3pt}
    \begin{tabular}{l|ccc|ccc|cc}
    \toprule
    \multirow{2}{*}{Methods} & \multicolumn{3}{c|}{RefCOCO} & \multicolumn{3}{c|}{RefCOCO+} & \multicolumn{2}{c}{RefCOCOg} \\
    & val & test-A & test-B & val & test-A & test-B & val-u & test-u \\ 
    \midrule
    \multicolumn{9}{c}{Expert Models}\\
    \midrule
    \color{gray}UNINEXT {\cite{UNINEXT}} & \color{gray} 92.6 & \color{gray} 94.3 & \color{gray} \underline{91.5} & \color{gray} 85.2 & \color{gray} 89.6 & \color{gray} 79.8 & \color{gray} 88.7 & \color{gray} 89.4 \\
    \color{gray}MDETR {\cite{kamath2021mdetr}} & \color{gray} 86.8 & \color{gray} 89.6 & \color{gray} 81.4 & \color{gray} 79.5 & \color{gray} 84.1 & \color{gray} 70.6 & \color{gray} 81.6 & \color{gray} 80.9 \\
    \color{gray}G-DINO-L {\cite{liu2023grounding}} & \color{gray} 90.6 & \color{gray} 93.2 & \color{gray} 88.2 & \color{gray} 82.8 & \color{gray} 89.0 & \color{gray} 75.9 & \color{gray} 86.1 & \color{gray} 87.0 \\
    \midrule
    \multicolumn{9}{c}{Large Multimodal Models}\\
    \midrule
    VistaLLM-7B {\cite{pramanick2024jack}} & 88.1 & 91.5 & 83.0 & 82.9 & 89.8 & 74.8 & 83.6 & 84.4  \\
    Ferret v2-7B {\cite{you2024ferret}}  & \underline{92.8} & \underline{94.7} & 88.7 & 87.4 & {92.8} & 79.3 & 89.4 & 89.3  \\
    CogCom-17B {\cite{qi2024cogcom}} & 92.3 & 94.6 & {89.2} & \textbf{88.2} & \textbf{92.8} & \underline{82.8} & 89.3 & \textbf{90.5} \\
    Visual CoT-7B {\cite{shao2024visualcot}}& 91.8 & 94.3 & 87.5 & 87.5 & 92.1 & 81.2 & 88.4 & 88.4 \\
    \midrule
    Griffon-R-9B & \textbf{93.2} & \textbf{95.1} & \textbf{90.2} & \textbf{88.4} & \underline{92.3} & \textbf{82.8} &  \textbf{89.8} & \underline{89.7} \\
    \bottomrule
    \end{tabular}
    }
\end{wrapfigure}

\textbf{Discussion on Understanding Quality.} As we have demonstrated in Sec. \ref{sec: mechanism}, understanding plays an important role in the whole mechanism for accurate visual reasoning. To quantitatively indicate the understanding quality, we choose the performance of Referring Expression Comprehension (REC) \cite{yu2016modeling, nagaraja2016modeling} as the metric, which incorporates both visual localization and attribute perception that are commonly used intrinsic capabilities during understanding for lots of reasoning tasks. As shown in Tab. \ref{table: rec}, our model outperforms the SOTA method Ferret v2 in this task and also visual reasoning methods CogCom and Visual CoT. The result verifies that our model can achieve accurate understanding with intrinsic capabilities and it further facilitates the precise visual reasoning of Griffon-R.

\begin{figure}[t]
  \centering
   \includegraphics[width=\linewidth]{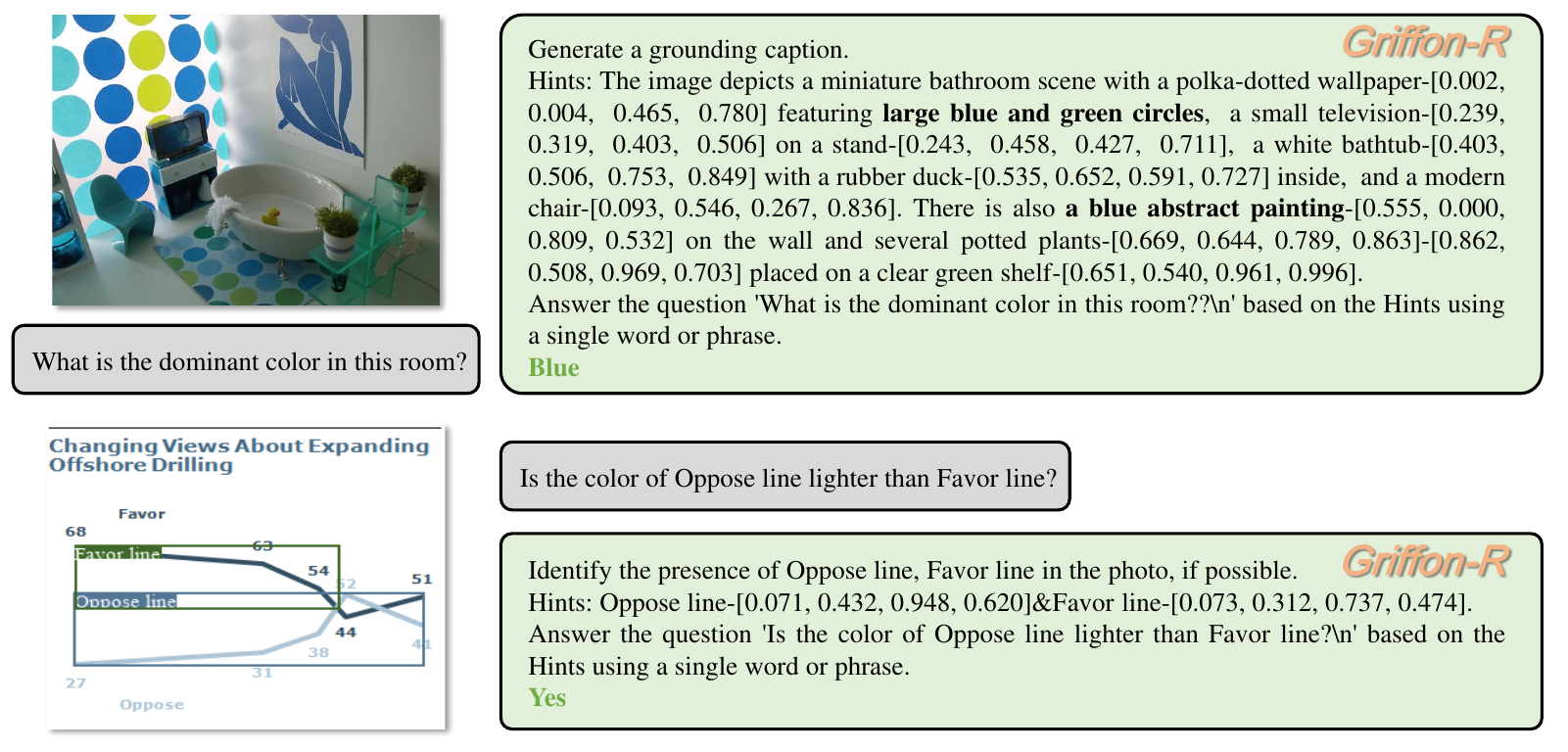}
   \caption{Visualization of Griffon-R's reasoning results. Correct answers are highlighted in bold green, and the relevant information within the long text leading to the answer is bolded.}
   \label{fig:vis}
\end{figure}

\textbf{Discussion on the unified visual reasoning mechanism.} To validate the effectiveness of our unified visual reasoning mechanism, we compare its accuracy and inference time to the toolkit-based SEAL-7B method.We evaluate with the V-Star$_{Spat.}$ benchmarks and static the average inference time per-sample. As shown in Tab. \ref{tab:mechanism}, in the compositional reasoning task of V-Star, the designed mechanism outperforms the toolkit-based pipeline by 1.3 points and is 13x faster in average inference time. These results justify that our mechanism not only boost visual reasoning capabilities without relying on visual expert modules but also delivers high efficinet inference.

\textbf{Discussion on training with the curated data annotations.} 
In this section, we validate the necessity of training with curated annotations for our mechanism. The first line (baseline) and the second line results are trained on the same data with Griffon-R using the raw annotations. The second line additionally follows our mechanism but requires the model to answer step-by-step. We evaluate these methods on the V-Star$_{Spat.}$ complex visual reasoning benchmark and the POPE object existence hallucination benchmark. As shown in Tab. \ref{tab:training}, simply providing visual cues (using UVRM) improves performance on tasks like POPE, which do not require compositional reasoning, by compensating for insufficient understanding. However, to achieve accurate visual reasoning, the model requires training on carefully curated datasets that enable it to understand the image, identify relevant cues, and reason based on different-pattern cues comprehension. This comparison emphasizes the effectiveness of our data-supported and unified visual reasoning mechanism design.
\begin{table}[]

    \centering
    \setlength{\tabcolsep}{3pt} 
    \begin{minipage}{0.41\linewidth}
        \centering
        \small
        \caption{Ablation study on the unified visual reasoning mechanism and toolkit-based mechanism.}
        \begin{tabular}{c|cc}
        \toprule
       Method  & V-Star$_{Spat.}$ &  Time/Sample\\
        \midrule
            Unified(ours)  &77.6& 0.336s  \\
            Toolkit-Based & 76.3  & 4.586s \\
        \bottomrule
        \end{tabular}
        \label{tab:mechanism}
    \end{minipage}%
    \hfill
    \begin{minipage}{0.55\linewidth}
        \centering
        \small
        \caption{Ablation study on the training and annotated data. UVRM denotes our mechanism.}
            \begin{tabular}{cc|cc}
            \toprule
            UVRM & 334k Ann.  & V-Star$_{Spat.}$ & POPE  \\
            \midrule
               &  & 75.0 & 89.1 \\
                $\surd$&  & 75.0 & 89.4 (+0.3) \\
                $\surd$&$\surd$ & 77.6 (+2.6)&  89.3 (+0.2)\\
            \bottomrule
        \end{tabular}
        \label{tab:training}
    \end{minipage}
\end{table}
\subsection{Visualization Results}
We display the qualitative performance of Griffon-R through Fig. \ref{fig:vis}. The results highlight Griffon-R effectively handles compositional visual reasoning across various scenes by thoroughly understanding the problem and visual cues, then performing context-based reasoning to generate the final answers.

\section{Conclusion}
In this paper, we present the unified visual reasoning mechanism that empowers LMMs to tackle compositional reasoning tasks end-to-end, without external expert models or toolkits. This unified mechanism introduces a human-like understand-think-answer process to reason based on individual questions flexibly instead of utilizing fixed paths and completes all steps in a single pass forward without multiple inferences. Moreover, we design a semi-automatic expert-supervised data generation engine to produce high-quality visual reasoning data corresponding to the design mechanism. We collect public data related to visual reasoning and re-annotate them with our designed data generation engine, and curate 334K visual instruction samples. Based on the curated data and designed mechanism, we present Griffon-R. Griffon-R achieves advancing results on visual reasoning benchmarks, including VSR and CLEVR, and also demonstrates comprehensive multimodal capabilities. We hope our attempts at a unified visual reasoning mechanism will facilitate the deep exploration in this field to achieve more general LMMs. In the Appendix, we provide a detailed discussion of our method’s limitations and its broader impact. In future work, we plan to explore additional reasoning paradigms and more diverse data to extend the applicability of our approach.

{
    \small
    \bibliographystyle{ieeenat_fullname}
    \bibliography{main}
}

\endgroup

\newpage
\appendix

\clearpage
\setcounter{page}{1}

\begin{center}
    \fontsize{20pt}{\baselineskip}\selectfont
    \textbf{Appendix}
    \vspace{0.4cm}
\end{center}

\tableofcontents
\newpage

\section{Details of Expert-Supervised Data Engine}

\label{app: data}
In this section, we illustrate the details of data curation process with our proposed semi-automatic expert supervised data engine of Sec. \ref{sec: data}. We start with the raw data collection, and then the instructions for the progressive annotation with AI expert.

\subsection{Raw Data Curation}
As described in Sec. \ref{sec: data}, we first collect the widely used diverse types of data that related to visual reasoning following the previous practice \cite{wu2024v, li2024vocot, shao2024visualcot} and curate a total of 334K visual-reasoning-oriented data using the proposed Expert-Supervised Data Generation Engine. These data are composed of two main parts: VQA-based data and caption data tailored to visual reasoning tasks.

\begin{table}[h]
    \centering
    \small
    \caption{Annotation sources of the 334K visual-reasoning data.}
    \adjustbox{max width=0.95\linewidth}{
    \begin{tabular}{c|cl}
        \toprule
        Type & Num. & Source \\
        \midrule
         VQA & 207K & 
             GQA\cite{gqa}, VAW \cite{pham2021learning}, VizWiz \cite{bigham2010vizwiz}, ChartQA \cite{masry-etal-2022-chartqa}, 
                DUE\_Benchmark \cite{borchmann2021due}, TextVQA \cite{textvqa} \\
         \midrule
         Instruction & 76K & LLaVA \cite{liu2024visual}, ALLaVA \cite{chen2024allava}, LVIS-Instruct4V \cite{wang2023instruct4v} \\
         \midrule
         Caption & 51K & ShareGPT-4V \cite{chen2023sharegpt4v}\\
         \bottomrule
    \end{tabular}
    }
    
    \label{tab: data distribution}
\end{table}

\textbf{VQA-Based Data.} The VQA-based data source from the general VQA datasets, instruction-following data, and text-oriented VQA datasets. We extract visual reasoning data based on two criteria: (1) Yes/No Answers: We select questions with ``yes'' or ``no'' answers, as these often involve tasks like comparisons or attribute judgments (e.g., color differentiation). (2) Reasoning Keywords: Using the relationship and attribute keywords appearing in GQA \cite{gqa}, VAW\cite{pham2021learning}, and DUE\_Benchmark \cite{borchmann2021due}, which we extract with the raw keywords in the annotations or with spaCy \cite{spacy} (\textit{eg.}, ``left'', ``right'', ``over'', ``red'', and ``plastic''), we identify Q\&A pairs that included these keywords. As summarized in Tab. \ref{tab: data distribution}, this process yielded 207K data from VQA annotations and 76K entries from instruction-based dialogues. For VAW data, we specifically utilize images aligned with GQA to generate more diverse visual reasoning questions. After acquiring these visual-reasoning-oriented VQAs, we follow the proposed data generation method to annotate these data.

\begin{table}[h]
    \centering
    
    \small
    \caption{Instruction templates for the AI expert.}
    \adjustbox{max width=0.95\linewidth}{
    \begin{tabular}{c|l}
    \toprule
        Type & Template\\
        \midrule
       I.1  &  \begin{tabular}{l}
            For the question:'\{question\}’, identify and focus on the specific physical objects or entities relevant to 
            answering it. \\Directly output the names of these objects or short descriptive phrases with key attributes. If the object is mentioned \\in the question, use the exact word from the question. If the object is not 
            explicitly mentioned, describe it concisely\\ using your own terms.
       \end{tabular}\\
       \midrule
       I.2  & \begin{tabular}{l}For the question:'\{question\}’, based on the context, identify the task used to gather the key entities 
       information \\like [\textit{TASKs}], if the question is too abstract, respond with ``Global Understanding''.\end{tabular}\\
       \bottomrule
    \end{tabular}
    }
    
    \label{tab: ann}
\end{table}

\textbf{Visual-Reasoning-Oriented Caption Data.} Caption data provide global context and are essential for the in-depth understanding of images. To make them suitable for visual reasoning, we apply the proposed data generation method, focusing first on identifying the key objects in the image that are critical for reasoning tasks. The annotation process involved checking whether these key objects are included, fixing them if missing, and manually labeling their bounding boxes. We utilized the detailed descriptions generated by ShareGPT-4V \cite{chen2023sharegpt4v}. After the above processing, we obtain the final expert-supervised visual-reasoning-oriented data, as listed in Tab. \ref{tab: data distribution}.

\subsection{AI Expert Annotation Details}
At the beginning of data generation process, we first utilize the AI expert Qwen2-VL-72B \cite{wang2024qwen2} to identify how to solve this problem with key objects or information listed (I.1) and plan how to gather these key information with intrinsic capabilities (I.2). We list the instructions to prompt the expert to finish these two tasks in Tab. \ref{tab: ann} respectively. Also, as we have mentioned in Sec. \ref{sec: data}, for the task annotation process that the AI expert is skilled at, we use the AI Expert to finish the annotation first. We mainly apply this strategy in tasks that require a global understanding, including caption and grounded caption. We directly used the instruction examples in the paper\cite{wang2024qwen2}. For generating this intermediate information at this stage, we perform batch inference using 8 NVIDIA A800 GPUs.

\begin{figure}[t]
    \small
  \centering
   \includegraphics[width=0.85\linewidth]{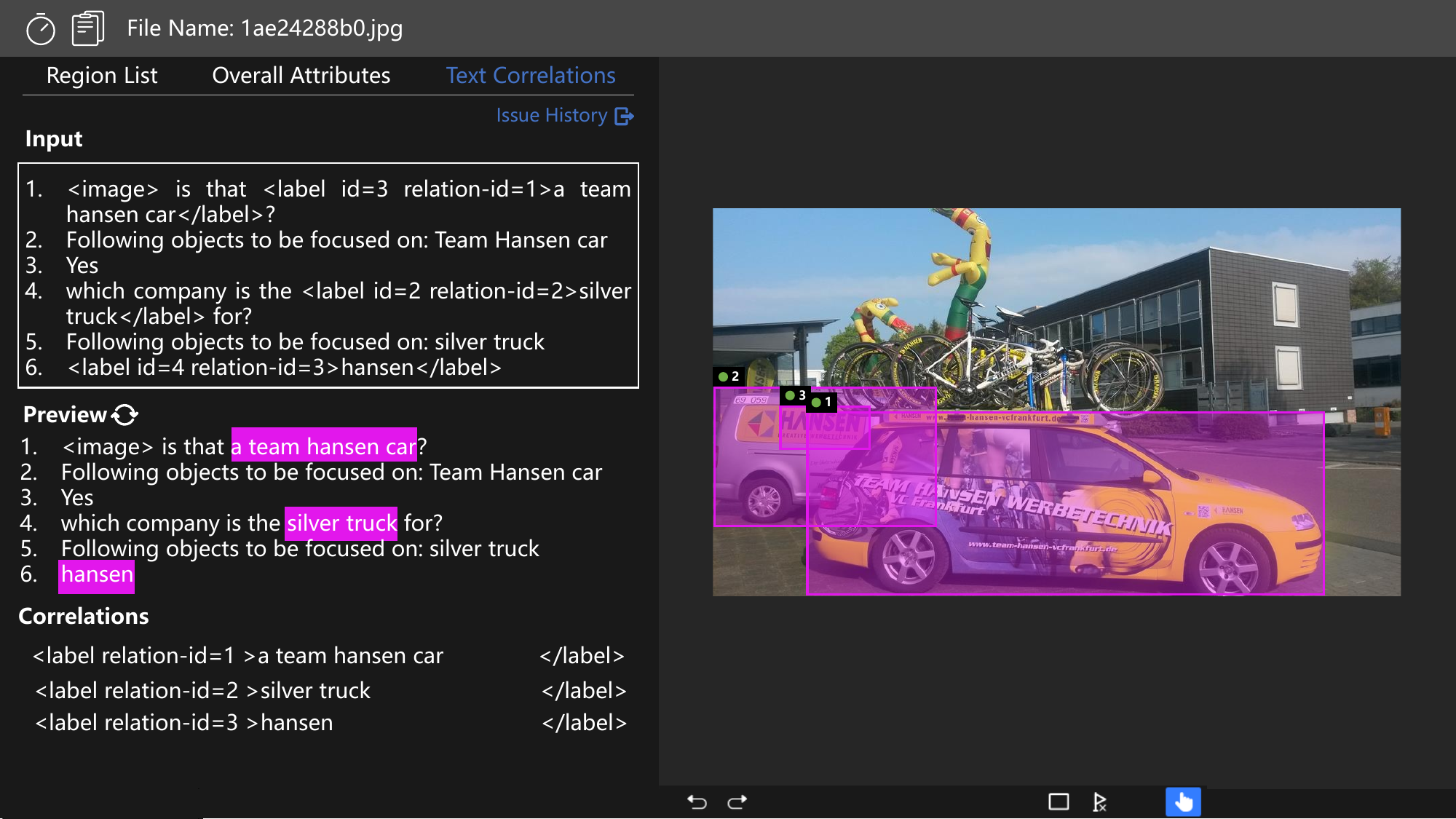}
   \caption{An example UI screenshot showcasing the annotation process for a question-answer pair requiring grounding. After reviewing and confirming the understanding process, the human verifying expert highlights the text based on the knowledge generated by the AI expert and annotates the corresponding bounding box by drawing on the image for each object.}
   \label{fig:UI}
\end{figure}
\subsection{Visualization of Human Expert Annotation}
As we have introduced in Sec. \ref{sec: data}, initial high-quality annotation is important. Therefore, we hire ten human annotation experts to help us. We provide a screenshot example of human expert annotating platform in Fig. \ref{fig:UI}.

\subsection{Discussion on Data Scaling}
Several works \cite{chen2023sharegpt4v, yuan2024osprey} have highlighted that it's more effective to scale up the annotated data based on the initial high-quality data. Therefore, we follow this insight to employ the human expert for annotation as existing advanced LMMs still fall short in tasks like multi-object visual grounding. For the visual reasoning data scaling up, we provide a brief discussion here. From the perspective of data amount, it's implementable to specifically train an expert model\cite{chen2023sharegpt4v} or our model using the curated data to further annotate a large amount of data. Then, a large expert-level model like ChatGPT\cite{chatgpt2024} or human experts can be used for checking. While for the scene scaling, when considering more scenes, it's possible to include more capabilities into our set and curated related data to support scenes like math, accounting, \textit{etc.} In this paper, we mainly focus on general natural scenes leveraging intrinsic capabilities like REG, caption, and visual grounding, and curate the 334K visual reasoning data. We will follow the discussion to scale up the data and generalize our mechanism to more scenes.

\section{Training Details}

\subsection{Training Data}
\label{app: train}
We list the training data used for the three training stages in Tab. \ref{tab: anndetail}, with the data volume and dataset type specified. As we have introduced in Sec. \ref{sec: model}, we employ the direct supervised fine-tuning (SFT) training strategy to build the mechanism within the Griffon-R. Therefore, we mix the visual reasoning data with general SFT data. While for the overlapping data with the visual reasoning data, we directly remove them.

\begin{table}[h]
    \centering
    \caption{Training data used in each stage. The Lang. represents the language-only instructions, the Inst. represents the vision-language instructions, the Gen. represents the general, the Text. represents text-oriented, and the Perc. represents the perception. Perceptions data include REC, REG, visual grounding and object detection, while VR stands for visual reasoning.}
    \small
    \adjustbox{max width=\linewidth}{
    \begin{tabular}{c|cc|l}
    \toprule
        Stage & Vol. & Type & Training Data\\
        \midrule
        \MakeUppercase{\romannumeral 1} & 1.2M & Caption & ShareGPT-4V \cite{chen2023sharegpt4v}\\
        \midrule
        \multirow{4}{*}{\MakeUppercase{\romannumeral 2}} & \multirow{4}{*}{3.0M} & REC & \begin{tabular}{@{}l@{}}RefCOCO/+\cite{yu2016modeling}, RefCOCOg\cite{nagaraja2016modeling}, GRefCOCO\cite{GREC}\end{tabular}\\
        \cmidrule{3-4}
        &&REG&\begin{tabular}{@{}l@{}}
             RefCOCO/+\cite{yu2016modeling}, RefCOCOg\cite{nagaraja2016modeling}, GRefCOCO\cite{GREC}, Flickr30K Entities\cite{flickr30k}, Visual Genome\cite{visualgenemo},\\
             Osprey\cite{yuan2024osprey}
        \end{tabular}\\
        \cmidrule{3-4}
        &&DET&\begin{tabular}{@{}l@{}}Objects365\cite{objects}, MSCOCO\cite{chen2015microsoft}, V3Det\cite{wang2023v3det}, Visual Genome\cite{visualgenemo}\end{tabular}\\
        \midrule
        \multirow{13}{*}{\MakeUppercase{\romannumeral 3}} & \multirow{13}{*}{3.9M} & Lang. & \begin{tabular}{@{}l@{}}  UltraChat\cite{ding2023enhancing}, Flan-mini\cite{ghosal2023flacuna}, OpenOrca\cite{OpenOrca}, MetaMathQA\cite{yu2023metamath}, ShareGPT,  MathInstruct\cite{yue2023mammoth}, \\WizardCoder\cite{luo2023wizardcoder} \end{tabular}\\
        \cmidrule{3-4}
        && Inst. & \begin{tabular}{@{}l@{}}LLaVA\cite{liu2024visual}, ALLaVA\cite{chen2024allava}, LVIS-Instruct4V\cite{wang2023instruct4v}\end{tabular}\\
        \cmidrule{3-4}
        && Caption & \begin{tabular}{@{}l@{}}ShareGPT4V\cite{chen2023sharegpt4v}, TextCaps\cite{sidorov2020textcaps}
        \end{tabular}\\
        \cmidrule{3-4}
        && Gen. VQA & \begin{tabular}{@{}l@{}}VQA v2\cite{vqa}, GQA\cite{gqa}, OK-VQA\cite{marino2019ok}, A-OKVQA\cite{schwenk2022okvqa}, SQA\cite{sqa}, VizWiz\cite{bigham2010vizwiz} \end{tabular}\\
        \cmidrule{3-4}
       && Text. VQA & \begin{tabular}{@{}l@{}}
           TextVQA\cite{textvqa}, OCR-VQA\cite{ocrvqa}, AI2D\cite{ai2d}, Synthdog\cite{kim2022donut},
           DVQA\cite{kafle2018dvqa}, ChartQA\cite{masry-etal-2022-chartqa}, DocVQA\cite{mathew2020docvqa}, \\InfoVQA\cite{mathew2022infographicvqa},
           DeepForm\cite{borchmann2021due}, KLC\cite{borchmann2021due}, WTQ\cite{borchmann2021due}, TabFact\cite{borchmann2021due}
       \end{tabular}\\
       \cmidrule{3-4}
       && Perc. & \begin{tabular}{@{}l@{}} RefCOCO/+\cite{yu2016modeling}, RefCOCOg\cite{nagaraja2016modeling}, GRefCOCO\cite{GREC}, Flickr30K Entities\cite{flickr30k}, Visual Genome\cite{visualgenemo},\\ Osprey\cite{yuan2024osprey},
             Objects365\cite{objects}, MSCOCO\cite{chen2015microsoft}, V3Det\cite{wang2023v3det}\\\end{tabular}\\
             \cmidrule{3-4}
        && VR & \textbf{Curated Visual Reasoning Data}\\
       \bottomrule
    \end{tabular}}
    
    \label{tab: anndetail}
\end{table}

\subsection{Trainable Parameter Setting}
In the first stage, we freeze the visual encoder and the LLM and leave the projector trainable. Then, we pretrain the whole model in stage \MakeUppercase{\romannumeral 2} and further finetune the whole model in stage \MakeUppercase{\romannumeral 3}.

\section{Limitations}
In complex scenarios, when the question-related objects are associated, the model's output sequence will grow, which can lead to an increase in the model's response time.

Our data pipeline incorporates Qwen2-VL-72B. As such, it inherits limitations common to MLLMs, including potential inaccuracies and the propagation of misinformation, although we have taken steps to reduce these through human verification. Concerning data usage, we explicitly state that Qwen's terms of use must be followed. The data may be used freely for research purposes; however, its use for other applications is restricted or necessitates a further request for permission.

\section{Broader Impact}
This paper presents work whose goal is to advance the field of MLLMs. There are many potential societal consequences of our work following the MLLMs, none of which we feel must be specifically highlighted here.

\end{document}